\pdfoutput=1

\documentclass[11pt]{article}

\usepackage{acl}

\usepackage{times}
\usepackage{latexsym}

\usepackage[T1]{fontenc}

\usepackage[utf8]{inputenc}
\usepackage{textcomp}
\DeclareUnicodeCharacter{25BC}{\textdownarrow}

\usepackage{microtype}

\usepackage{inconsolata}

\usepackage{booktabs}
\usepackage[table]{xcolor} 
\usepackage{enumitem}

\usepackage[normalem]{ulem}

\usepackage{amssymb}

\usepackage{graphicx}
\usepackage{makecell}
\usepackage{bm}

\newcommand{\CR}[1]{#1}

\makeatletter
\newsavebox{\saved@arstrutbox}
\newcommand{\smallerow}[2][.75]{%
  \noalign{%
    \global\savebox{\saved@arstrutbox}{\copy\@arstrutbox}%
    \dimen0=#1\ht\@arstrutbox
    \dimen2=#1\dp\@arstrutbox
    \global\setbox\@arstrutbox\hbox{\vrule height \dimen0 depth \dimen2 width 0pt}%
  }%
  {\scriptsize #2}%
  \noalign{\global\setbox\@arstrutbox\copy\saved@arstrutbox}%
}
\makeatother

\title{Assessing the Impact of Typological Features on \\ Multilingual Machine Translation in the Age of Large Language Models}

\author{
Vitalii Hirak\textsuperscript{1}, Jaap Jumelet\textsuperscript{2}, Arianna Bisazza\textsuperscript{2} \\
\textsuperscript{1}Data \& Knowledge Engineering, Heinrich Heine University \\
\textsuperscript{2}Center for Language and Cognition (CLCG), University of Groningen \\
\texttt{vitalii.hirak@hhu.de \{j.w.d.jumelet, a.bisazza\}@rug.nl}
}

\begin{document}
\maketitle

\begin{abstract}
Despite major advances in multilingual modeling, large quality disparities persist across languages. Besides the obvious impact of uneven training resources, typological properties have also been proposed to determine the intrinsic difficulty of modeling a language. The existing evidence, however, is mostly based on small monolingual language models or bilingual translation models trained from scratch. We expand on this line of work by analyzing two large pre-trained multilingual translation models, NLLB-200 and Tower+, which are state-of-the-art representatives of encoder-decoder and decoder-only machine translation, respectively. Based on a broad set of languages, we find that target language typology drives translation quality of \CR{both models}, even after controlling for more trivial factors, such as data resourcedness and writing script. Additionally, languages with certain typological properties benefit more from a wider search of the output space, suggesting that such languages could profit from alternative decoding strategies beyond the standard left-to-right beam search. To facilitate further research in this area, we release a set of fine-grained typological properties for 212 languages of the FLORES+ MT evaluation benchmark.
\end{abstract}

\section{Introduction} \label{intro}

Despite major advances in multilingual modeling, the quality of language technologies still varies widely across languages \cite{joshi-etal-2020-state, blasi-etal-2022-systematic, sarti-etal-2022-divemt}.
These inequalities are largely due to the uneven availability of training data. However, some languages also appear to be intrinsically more difficult to model than others by modern approaches, a variability that has been connected to typological properties by a rich line of work \cite{birch-etal-2008-predicting, cotterell-etal-2018-languages,mielke-etal-2019-kind,bugliarello-etal-2020-easier,bisazza-etal-2021-difficulty,park-etal-2021-morphology,arnett-bergen-2025-language}.

To isolate intrinsic modeling difficulty from other factors, those studies strongly prioritize the comparability of training corpora. While principled, this choice constrains evaluations to a very small set of existing multi-parallel datasets, typically, Europarl~\citep{koehn-2005-europarl} or the Bible \citep{mayer-cysouw-2014-creating}, which are limited in typological diversity or in size and domain.
Moreover, models trained from scratch on such corpora are poor representatives of current practices in MT and language modeling in general, leaving the open question: can typological properties explain state-of-the-art MT quality in the age of LLMs?

To provide an answer, we expand on previous work by evaluating two large pre-trained multilingual models in a wide, typologically diverse set of languages, while using approximations of language resourcedness to control for data size effects. Additionally, we explore how widening the search for a high-probability sequence during inference affects translation quality in languages of different typology, calling into question the optimality of using a single decoding strategy across many different languages. With a focus on word order and morphological complexity,
we consider a broad set of fine-grained features, following recent trends in typology where continuous (or gradient) features are increasingly preferred over coarse-grained categorical ones \citep{levshina2023we, baylor2024multilingualgradientwordordertypology}.
Our work makes the following contributions:

\begin{itemize}
    \item Identifying specific typological features that predict translation quality of two widely used multilingual MT models --- the encoder-decoder NLLB-200 model \citep{nllbteam2022language} and the decoder-only Tower+ model \citep{towerplus} --- across a total of 7 source and 124 target languages.
    \item Analyzing the interplay between optimal beam size and typological properties of the generated language.
    \item Compiling a dataset of fine-grained, continuous typological features for the 212 target languages in the FLORES+ benchmark \citep{nllbteam2022language} to facilitate further research on language-specific decoding strategies.\footnote{\href{https://github.com/v-hirak/explaining-MT-difficulty}{\tt github.com/v-hirak/explaining-MT-difficulty.}}
\end{itemize}

\noindent
Leveraging a typologically diverse selection of languages, we show that target language typology drives translation quality of the NLLB-200 model, even after accounting for more trivial factors, such as data resourcedness and writing script. 
Additionally, we uncover a large variability in optimal beam size across target languages, and find that widening beam search yields significantly higher-probability outputs for languages of certain typologies. Important disparities in translation performance are also observed in the LLM-based Tower+ model, \CR{although across fewer significant factors}, calling for further experimentation with this type of models.


\section{Background} \label{background}

\subsection{Typological Language Properties} 
\label{background:typology}

Languages differ from one another in various aspects such as phonology, morphology, and syntax. One approach to measure this variation is to assign \textit{discrete} categorical values to languages, as exemplified by the World Atlas of Language Structures (WALS) \citep{wals} and GramBank \citep{skirgaard2023grambank}. Recent literature argues, however, that using \textit{continuous} languages properties more accurately reflects the variability of natural languages and is more appropriate in the realm of NLP \cite{levshina2023we, baylor2024multilingualgradientwordordertypology}. Continuous approaches towards measuring the morphological complexity and word order flexibility range from simple ratios \cite{chotlos1944iv, xanthos2011role} to applying information-theoretic principles \cite{juola1998measuring, bentz2016wordentropynaturallanguages} to using the accuracy of machine learning (ML) models on the task of predicting inflected word forms \cite{cotterell2019complexity}.
In line with these findings, we use a combination of continuous language properties computed in prior work as well as properties calculated ourselves on the basis of the FLORES+ dataset.

\subsection{Typology and Modeling Difficulty} \label{background:modeling}
A number of studies have measured and tried to explain the intrinsic difficulty of modeling different languages using controlled setups, that is with fixed size and, where possible, comparable content of training data.
\citet{cotterell-etal-2018-languages} and \citet{mielke-etal-2019-kind} study monolingual LSTM models trained on the Europarl corpus. They conclude that only general statistical properties, like raw character sequence length and the tokenization-specific word inventory size, correlate significantly with LM surprisal values. Their considered WALS features and other language properties do not. \citet{park-etal-2021-morphology} revisit the question including more languages (from the Bible corpus) and more morphological features from WALS. 
They find various morphological features to correlate significantly with LM difficulty for a BPE-based model, but not for a character-level one.
\citet{arnett-bergen-2025-language} analyse 22 monolingual Transformer models from the Goldfish suite \citep{chang2024goldfish} and find mismatches in data size calculations to better predict perplexity differences compared to morphological complexity and tokenizer quality. 

In the translation domain, target language morphological complexity and language relatedness were found to be significant predictors of translation performance already in the era of pre-neural statistical MT \cite{birch-etal-2008-predicting}.
More recently, \citet{bugliarello-etal-2020-easier} train Transformer models on Europarl language pairs and find only type/token ratio (TTR) to correlate significantly with translation difficulty.
\citet{wan2022fairness} denies the role of morphological complexity in driving modeling difficulty and instead attributes disparities to mismatches in representational granularity among languages (e.g. longer encodings for some writing scripts).
\citet{bisazza-etal-2021-difficulty} use synthetic versions of English to investigate whether languages with flexible word order and case marking are more difficult to translate by NMT models. They find that, in low-resource settings, such languages are harder to learn than their fixed-order, no-marking counterparts.
Concurrently to this work, \citet{ploeger-etal-2025-cross} construct a multi-parallel dataset and use it to train bilingual NMT models and analyze translation difficulty in the scope of 8 European languages. They find that NMT difficulty varies across language pairs and can be predicted based on their genetic and syntactic similarity. While being highly controlled, their experimental setup is not necessarily representative of state-of-the-art NMT and is very limited in typological diversity.

Another set of studies attempts to uncover the inductive biases of language models by training (and evaluating) them on artificial languages of different typologies \citep{white-cotterell-2021-examining,kallini-etal-2024-mission,kuribayashi-etal-2024-emergent,el-naggar-etal-2025-gcg,yang-etal-2025-anything}.

We depart from these lines of work by studying considerably larger, pre-trained multilingual models representative of state-of-the-art MT, at the cost of control over the training data.
A similarly pragmatic approach is taken by \citet{arnett-bergen-2025-language} in their analysis of multilingual models in various tasks (but not MT). Their hypotheses were later reassessed in \citet{poelman-etal-2025-confounding}, who identified confounding factors in the experimental setup that should be considered, such as languages considered, tokenization algorithms, training data characteristics, and performance indicators.
Also related to our work, \citet{sarti-etal-2022-divemt} assess the usefulness of Google Translate and mBART-50 translations by a human post-editing study. Their chosen target languages are typologically diverse, but are too few (six) to draw any reliable conclusions on the effect of typological properties.

\subsection{Decoding Algorithms} \label{background:beam-search}

Decoding or generation algorithms serve to search for a high-probability sequence through an intractably large output space. One of the most widely used algorithms is \textit{beam search} \cite{wu2016google, brown2020languagemodelsfewshotlearners, costa2022no, raffel2023exploringlimitstransferlearning}, a simple deterministic approximation to maximum a-posteriori decoding, where a `beam' of the $k$ most probable partial hypotheses is kept at each time step. Despite its simplicity, beam search with a small $k$ value (e.g. 3 to 5) typically yields considerably higher-probability outputs and better task performance than greedy decoding (a special case of beam search with $k=1$) at the cost of slower inference, across tasks and settings 
\citep{junczys-dowmunt-etal-2016-neural,freitag-al-onaizan-2017-beam,kulikov-etal-2019-importance, park2020decoding}.
Output quality has been found to deteriorate beyond a certain beam size \citep{koehn-knowles-2017-six,cohen2019empirical}.
While all these works consider very few, mostly high-resource languages,
our evaluation covers a much broader and diverse set of language pairs.

Beyond beam search, many other decoding algorithms exist, including deterministic
and stochastic ones (see \citet{welleck2024from} for an extensive survey).
Given common practice in the MT community as well as recent evidence on the optimality of beam search for translation in state-of-the-art LLMs \citep{shi-etal-2024-thorough}, we focus here on varying the width of beam search and leave the exploration of alternative algorithms to future work.

To our knowledge, no work has studied the interplay between optimal decoding strategy and the typological properties of the generated language.

\section{Experimental Setup} \label{setup}

This section describes our experimental setup, including the choice of models and datasets, languages, and evaluation metrics.

\paragraph{Translation Models}
We opt for two multilingual pretrained models: NLLB-200 \cite{nllbteam2022language} and Tower+ \cite{towerplus}. NLLB-200 3.3B\footnote{\href{https://huggingface.co/facebook/nllb-200-3.3B}{\tt hf.co/facebook/nllb-200-3.3B}.} is an encoder-decoder Transformer NMT model with 3.3 billion parameters, capable of translating among 202 languages. Tower+ 9B\footnote{\href{https://huggingface.co/Unbabel/Tower-Plus-9B}{\tt hf.co/Unbabel/Tower-Plus-9B}. For the prompt template used to generate translations, see App.~\ref{app:prompt}} is a decoder-only multilingual LLM post-trained for MT, and it was shown to often outperform larger general-purpose open-weight and proprietary models such as Llama 3.3 70B and GPT-4o in translation tasks \cite{towerplus}. While Tower+ was explicitly post-trained only on 22 languages, its underlying LLM model, Gemma2 9B \cite{gemmateam2024gemma2improvingopen}, exhibits multilingual capabilities that extend to more languages. 
Both NLLB and Tower models use a single subword vocabulary shared across languages, as is standard practice in modern multilingual models.

\paragraph{Evaluation Dataset}
To define the set of languages studied and for translation material, we use the wide-coverage multi-parallel FLORES+ machine translation evaluation benchmark dataset.\footnote{\href{https://huggingface.co/datasets/openlanguagedata/flores_plus}{\tt hf.co/datasets/openlanguagedata/flores\_plus}.} FLORES+ is composed of English sentences equally sampled from Wikinews, Wikijunior, and Wikivoyage and manually translated into over 200 languages. For our experiments, we use the \texttt{dev} split of the dataset comprising 997 sentences.

\paragraph{Translation Directions}
Since we are primarily interested in the effect of \textit{target} languages on translation difficulty and decoding requirements, we select 124 typologically diverse target languages from FLORES+, taking into account NLLB-200 language coverage, the availability of language properties, and computational resources at hand.\footnote{For the list of target languages, see Appendix \ref{app:tgt_langs}.} Our first set of experiments involves translating from English into a large variety of target languages. To ensure our results are not dependent on this specific source language choice, we then experiment with six additional source languages (Arabic, Italian, Dutch, Turkish, Ukrainian, and Vietnamese), which were selected in previous work \cite{sarti-etal-2022-divemt} to ensure typological diversity and comparable data resourcedness.

\paragraph{Beam Size}
In light of previous findings on the ineffectiveness of very wide beam search \cite{koehn-knowles-2017-six, cohen2019empirical}, we limit our selection of beam sizes used in generating translations to $k\in\{1,3,5,7\}$. 

\paragraph{Evaluation Metrics} \label{methodology:evaluation}
As our translation quality metric, we use chrF++ \cite{popovic-2015-chrf},\footnote{We use the sacreBLEU implementation \cite{post-2018-call}.} which is based on character-level n-gram overlap between the hypothesis and reference translations.
Compared to word-level metrics such as BLEU \citep{papineni-etal-2002-bleu}, chrF++ is better suited for languages with rich morphology. 
However, being based on surface-level matching, it can still fail to capture semantic similarities between MT output and reference.
Modern translation quality metrics addressing this issue, such as COMET \citep{rei-etal-2020-comet} and  MetricX \citep{juraska-etal-2023-metricx}, correlate better with human judgments. However, their reliance on pre-trained neural encoders or LLMs makes them unreliable for low-resource languages \citep{falcao-etal-2024-comet,singh-etal-2024-good,wang-etal-2024-evaluating,sindhujan-etal-2025-llms}.
Given our strong focus on cross-lingual comparability, we thus opt for a simpler, model-free metric.
Note that chrF++ is still used routinely as an additional metric in large-scale multilingual MT evaluations \citep{alves2024toweropenmultilinguallarge, towerplus} and for low-resource pairs \citep{kocmi-etal-2025-findings}.
Appendix~\ref{app:other metrics} reports overall scores of the NLLB model with alternative metrics (BLEU and COMET), showing similar trends by beam size on average.

\section{Language Properties} \label{setup:properties}

Here we outline our choice of language properties used to estimate the impact of a language typology on NMT (see App.~\ref{app:properties} for a more detailed description of each property).
Besides language resourcedness and coarse-grained source-target distances, we focus on features of morphological complexity and word order of the language being generated.

\subsection{Language Resourcedness} \label{setup:properties:size}

The proportion of a language in the model's training data is clearly an important factor for translation quality.
Unfortunately, this information is often unavailable for large pre-trained models. Even in the case of open-source models like NLLB-200, calculating a language proportion is complicated by the size of the dataset and absent or imprecise meta-data.\footnote{We initially attempted to reconstruct NLLB-200 training data size proportions by language, but due to important details missing (such as the amount of back-translated bitext), our estimates were ultimately inaccurate and unreliable.}

In light of this, we approximate the \textit{general resourcedness of a language} using language size data from the GlotCC broad-coverage CommonCrawl corpus \cite{glotcc}. Specifically, we collect content length values for 210 out of 212 languages in FLORES+.
While this is a very rough approximation of what our evaluated models were exposed to, we assume that large disparities within CommonCrawl will correlate overall with large disparities of language presence on the Web, which provides the large bulk of training data for modern translation systems.

\subsection{Source-Target Distances}

Following previous work on NMT \cite{sarti-etal-2022-divemt} and cross-lingual transfer \cite{lin-etal-2019-choosing}, we adopt the URIEL typological database \cite{littell-etal-2017-uriel} and query six types of typological distances pre-computed by aggregating broad categories of typological features: \textbf{genetic}, \textbf{geographic}, \textbf{syntactic}, \textbf{inventory}, \textbf{phonological}, and \textbf{overall features}  distance.\footnote{Concurrently to our work, \citet{goot-etal-2025-distals} released a toolkit providing a larger variety of language distances, which could be used to extend our analysis in future work.} Additionally, we experiment with a simple \textbf{same\_script} binary feature, capturing the advantage of language pairs having the same writing script (and, potentially, shared subwords).

\subsection{Morphological Complexity Measures} \label{setup:properties:morphology}

To estimate the morphological complexity of a language, we make use of eight continuous measures that were precalculated and made available by \citet{ccoltekin2023complexity}. These measures were computed on the basis of Universal Dependencies and are available for 33 languages of the FLORES+ dataset.
\textbf{Information in Word Structure } (\texttt{WS}) compares the information content (i.e. entropy) of the original text with its compressed version \cite{juola1998measuring}.
\textbf{Word and Lemma Entropy} (\texttt{WH}, \texttt{LH}): word entropy is based on word frequency distribution of a text \cite{bentz-etal-2016-comparison}; \citet{ccoltekin2023complexity} additionally calculate the entropy of lemmas.
\textbf{Mean Size of Paradigm} (\texttt{MSP}) is calculated by dividing the number of word forms in a text by the number of lemmas \cite{xanthos2011role}.
\textbf{Inflectional Synthesis} (\texttt{IS}) is the maximum number of inflection categories that can be expressed by a standalone verb.
\textbf{Morphological Feature Entropy} (\texttt{MFH}) reflects the usage of morphological features (e.g. grammatical cases) and their values.
\textbf{Inflection Accuracy} (\texttt{IA}) is the accuracy of an ML model on the task of predicting inflected forms given lemma and grammatical features.\footnote{For consistency with other metrics, we report results with \textit{negative} inflection accuracy \texttt{-IA}, such that higher \texttt{-IA} means \textit{lower} accuracy, reflecting higher complexity.}
Finally, we express the \textbf{Type/Token Ratio} --- the number of unique words divided by total words --- in three ways: \textbf{TTR} \citep{chotlos1944iv} (on both UD and FLORES+), \textit{Root Type/Token Ratio} (\textbf{RTTR}) \cite{guiraud1959problemes}, and \textit{Moving Average Type/Token Ratio} (\textbf{MATTR}), which is TTR calculated inside a sliding window \cite{covington2010cutting} (both computed on FLORES+).

\subsection{Gradient Word Order Measures} \label{setup:properties:word-order}

We include a number of gradient word order measures proposed and computed by \citet{levshina2019token} and \citet{levshina2023we}.
\textbf{Average word order entropy of dependents and codependents} ({\tt h\_dep}/{\tt h\_codep}) is the entropy of different word order patterns of \textit{dependencies} (e.g. verb-subject and noun-adposition relations) and \textit{codependencies} (e.g. subject and object of the same verb).
\textbf{Proportion of Subject-Object order} ({\tt SO\_prop}) is based on the frequencies of clauses where subject comes before object.\footnote{\CR{For consistency with other metrics, we report results with \textit{negative} SO order proportion {\tt -SO\_prop}, such that higher {\tt -SO\_prop} values reflect \textit{lower} predictability of word order.}}
\textbf{Percentage of head-final phrases} ({\tt head\_finality}) approximates the preference of a language towards head-initial or head-final phrases.


\section{Results} \label{results}

Having collected the set of language properties, we analyze the cross-lingual variability of translation quality and its improvement as a function of beam size, across a wide set of languages. 
We begin with a correlation study (\S\ref{results:nllb:quality}) to get a general idea of which typological properties of target languages correlate significantly with chrF++ scores of the NLLB-200 model. 
We then use these insights to narrow down our selection of language features and use them in a more focused regression setup (\S\ref{results:nllb:regression:chrf5}, \S\ref{results:nllb:regression:gain}), where we evaluate their effect on predicting translation difficulty and its change at a higher beam size. 
Finally, we extend the regression experiments to the Tower+ model to determine whether our findings generalize to decoder-only LLMs used nowadays for MT (\S\ref{results:tower}).

\subsection{Encoder-Decoder NLLB-200} \label{results:nllb}
Due to computational constraints, we conduct the full set of analyses on the lighter-weight NLLB-200 model, which also provides official support for a larger set of languages than Tower+.

\subsubsection{Translation Quality} \label{results:nllb:quality}
As a first exploration of language properties affecting translation performance, we measure the correlation between all our language properties (\S\ref{setup:properties}) and chrF++ scores for translations from English at beam size $k=5$, commonly used in practice. We report Spearman's rank correlation as property values are not normally distributed. 
Due to the varying availability of linguistic features, finding a subset of languages where all the features are specified would result in a very limited sample. 
We therefore compute correlations on subsets of varying sizes to leverage as many data points as possible.

\begin{figure}[t]
    \centering
    \includegraphics[width=\linewidth]{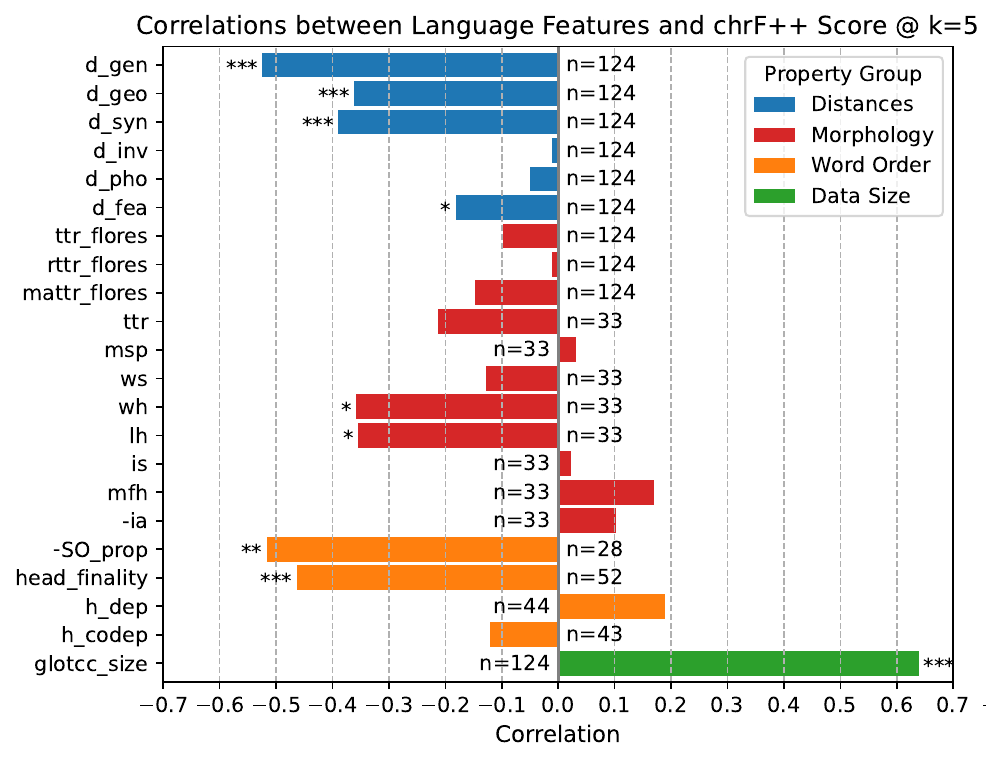}
    \caption{Spearman correlations between continuous language properties and NLLB-200 chrF++ translation quality scores (a character n-gram based translation quality metric, cf.\S{\ref{methodology:evaluation}}) at beam size $k=5$. Source language is English. Sample sizes (i.e. number of target languages) for each property are indicated next to their respective bars. Correlations significant at $p<0.05$ are marked with *, at $p<0.01$ with **, at $p<0.001$ with ***.}
    \label{fig:corr-chrf_5-nllb}
\end{figure}

As shown in Fig.~\ref{fig:corr-chrf_5-nllb}, various language properties correlate significantly with translation quality scores. 
The strong positive correlation with \textit{GlotCC size} confirms that these estimates can be used to approximate language proportions in our model's training data. The source-target typological distances also behave as expected: \textit{genetic}, \textit{geographic}, and \textit{syntactic} distances inversely correlate with chrF++, confirming that more disparate language pairs are harder to translate. Unsurprisingly, no correlation is observed for the phonological distances (\textit{phonological} and \textit{inventory}).
As for target morphological complexity, 2 out of 11 continuous measures correlate significantly with translation quality, namely \textit{word and lemma entropy} (\texttt{WH}, \texttt{LH}), which reflect the average information content of words. This implies that target languages packing more information into word structure (rather than sentence structure) tend to have lower translation quality in NLLB.
Finally, two target word order measures have some of the strongest correlations with translation quality: languages with less predictable word orders (higher \texttt{-SO\_prop})
and languages with a stronger preference for head-final phrases (higher \texttt{head\_finality}) are more challenging to translate into.

\subsubsection{Predicting chrF++ Scores}
\label{results:nllb:regression:chrf5}

\begin{table}
    \centering
    \footnotesize
    \setlength{\tabcolsep}{3.5pt}
    \begin{tabular}{lrrrrr}
        & \multicolumn{5}{c}{{\textit{Source language}: \textbf{All} ($n=364$)}} \\[3pt]
        \textbf{Feature} & $\bm{\beta}$ & \textbf{\textit{F}} & \textbf{\textit{p}-value} & \textbf{\textit{LR}} & \textbf{Adj. \textit{R\textsuperscript{2}}} \\
        \midrule
    
        \texttt{src\_language} & -- & 9.02 & $<$0.001 & -- & 0.07 \\
        {\scriptsize\hspace{0.5cm}\textit{Arabic}} & \scriptsize\textbf{-3.65} \\
        {\scriptsize\hspace{0.5cm}\textit{Dutch}} & \scriptsize\textbf{-10.3} \\
        {\scriptsize\hspace{0.5cm}\textit{Italian}} & \scriptsize\textbf{-8.48} \\
        {\scriptsize\hspace{0.5cm}\textit{Turkish}} & \scriptsize\textbf{-7.18} \\
        {\scriptsize\hspace{0.5cm}\textit{Ukrainian}} & \scriptsize\textbf{-5.16} \\
        {\scriptsize\hspace{0.5cm}\textit{Vietnamese}} & \scriptsize\textbf{-8.58} \\[2pt]
    
        \texttt{tgt:glotcc\_size} & \textbf{3.34} & 56.5 & $<$0.001 & 59.1$^{*}$ & 0.20 \\
        \texttt{is\_same\_script} & \textbf{5.67} & 26.8 & $<$0.001 & 63.6$^{*}$ & 0.33 \\
        \texttt{d\_gen} &\textbf{-1.24} & 6.2 & 0.01 & 8.75$^{*}$ & 0.34 \\
        \texttt{tgt:mattr} & -0.67 & 2.02 & 0.16 & 7.49$^{*}$ & 0.36 \\
        \texttt{tgt:head\_fin} & \textbf{-2.58} & 27.2 & $<$0.001 & 27.1$^{*}$ & 0.40 \\
        \midrule
    \end{tabular}
    \quad
    \caption{
    Regression results for predicting chrF++ scores at $k=5$ for NLLB-200 translations. 
    $\beta$ indicates the $\beta$-coefficients, \textit{F} indicates the $F$ scores of the ANOVA type II test, \textit{LR} --- the log-likelihood ratio of incrementally adding each variable (from top to bottom, significant ratios marked by *), and the adjusted $R^2$ --- the explained variance of the incremental linear model (top to bottom). Coefficient values ($\bm{\beta}$) marked in bold are significant ($p<0.05$).}
    \label{table:regression-chrf_5-nllb}
\end{table}

Based on the correlation results, we narrow down our selection of features and languages to allow for a more controlled linear regression setup.
Specifically, we predict chrF++ scores at $k=5$ including obvious factors such as the source language id,\footnote{The limited number of source languages (7) prevent us from properly modeling source language properties. Empirically, using src\_id or source features leads to a similar fit.} the log-transformed GlotCC estimate of the target language's presence in training data,\footnote{Because the values are positive and span orders of magnitude, we log-transform them to reduce skewness and heteroscedasticity prior to the regression analysis.} source-target language genetic distance, and whether the language pair shares the same script.\footnote{\CR{As suggested later by an anonymous reviewer, we additionally conducted regression experiments to assess the effect of \textit{token fertility} (average number of tokens per word) on chrF++ scores of both models. Ultimately, the effect emerged as non-significant after all other factors were accounted for.}} On top of that, we incorporate \textit{moving average type/token ratio} (MATTR) as a measure of target morphological complexity and target \textit{head-finality} as a word order measure.\footnote{We also added interactions with source language MATTR and head-finality, but results were insignificant.} This feature set ensures we have data for 52 target languages, and we additionally expand our source language selection to seven languages, resulting in a total of $7 \times 52 = 364$ data points.

Table \ref{table:regression-chrf_5-nllb} summarizes the regression results.
Overall, the significance and directionality of the effects on chrF++ follow our intuitions, barring target language MATTR. Namely, larger source-target genetic distance, prevalence of head-final phrases in the target language, and translating from source languages other than English all result in \textit{lower} translation quality. Importantly, these trends are observed after controlling for data size effects and script similarity.\footnote{We provide a closer look at the effects of head-finality and MATTR for individual language pairs in App.~\ref{app:regplots}.}

\subsubsection{Beam Width} \label{results:nllb:regression:gain}

\begin{table*}[ht]
    \centering
    \footnotesize
    \setlength{\tabcolsep}{6.5pt}
    \begin{tabular}{lrrrrrr}
        & \multicolumn{5}{r}{\textit{\textbf{Metric}}: $\Delta chrF_{1;7}$ ($n=364$)} \\[3pt]
        \textbf{Feature} & $\bm{\beta}$ & \textbf{\textit{F}} & \textbf{\textit{p}-value} & \textbf{\textit{LR}} & \textbf{Adj. \textit{R\textsuperscript{2}}} \\
        \midrule
        \texttt{src\_language} & \textbf{-0.3} & 4.40 & $<$0.001 & -- & 0.05 \\
        \texttt{tgt:glotcc\_size} & \textbf{0.15} & 17.2 & $<$0.001 & 14.2$^*$ & 0.09 \\
        \texttt{is\_same\_script} & -0.02 & 0.06 & 0.81 & 1.72 & 0.09 \\
        \texttt{d\_gen} & 0.05 & 1.48 & 0.22 & 2.00 & 0.09 \\
        \texttt{tgt:mattr} & 0.02 & 0.35 & 0.56 & 1.04 & 0.09 \\
        \texttt{tgt:head\_fin} & 0.06 & 2.38 & 0.12 & 2.45 & 0.09 \\
        \bottomrule
    \end{tabular}
    \quad
    \begin{tabular}{rrrrr}
    \multicolumn{5}{r}{\textit{\textbf{Metric}}: $\Delta prob_{1;7}$ ($n=364$)} \\[3pt]
     $\bm{\beta}$ & \textbf{\textit{F}} & \textbf{\textit{p}-value} & \textbf{\textit{LR}} & \textbf{Adj. \textit{R\textsuperscript{2}}} \\
    \midrule
    \textbf{0.18} & 3.6 & 0.002 & -- & 0.01 \\
    \textbf{-0.21} & 180.8 & $<$0.001 & 141.8$^*$ & 0.33 \\
    \textbf{-0.08} & 4.5 & 0.03 & 37.2$^*$ & 0.39 \\
    \textbf{0.09} & 26.9 & $<$0.001 & 29.7$^*$ & 0.44 \\
   \textbf{ 0.05} & 10.6 & 0.001 & 23.1$^*$ & 0.47 \\
    \textbf{0.12} & 46.3 & $<$0.001 & 45.0$^*$ & 0.53 \\
    \bottomrule
    \end{tabular}
    \caption{
    Regression results for predicting chrF++ gain (\textit{left}) and probability gain (\textit{right}) for NLLB-200 translations. The model is fitted on the results of all source languages. Coefficient values ($\bm{\beta}$) marked in bold are significant ($p<0.05$). Here $\bm{\beta}$ for source language denotes a mean effect of source language with English as a reference value (see App. \ref{app:regression-gain-nllb-src_lang} for effect breakdown by source language).
    }
    \label{table:regression-gain-nllb}
\end{table*}

\begin{figure}[t]
    \centering
\includegraphics[width=\linewidth]{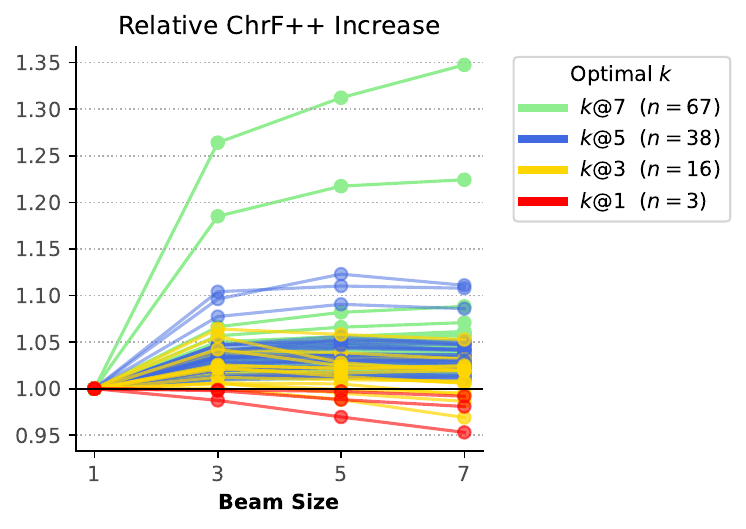}
    \caption{Relative increase in chrF++ for NLLB-200, translating from English to 124 different target languages. Curves are colored by their optimal beam size.}
    \label{fig:rel_chrf}
\end{figure}

We now shift our attention to the changes in translation quality as a function of beam size.
First, we plot beam size against the relative increase in chrF++ for the NLLB-200 model translating from English into 124 target languages (Fig.~\ref{fig:rel_chrf}). We find a large variation across languages, with roughly half of the languages benefiting from a wider beam size ($k=7$), suggesting that a language-specific choice of decoding strategy could be beneficial in terms of quality and efficiency.
Next, we investigate whether this variability can be partly explained by target typological properties. To this end, we define two measures of \textit{gain by beam size}:
\begin{itemize}
\item \textbf{chrF++ gain}: for each source-target language pair, we calculate the gain in chrF++ as
$\Delta chrF_{1;7} = chrF_7 - chrF_1$,
where $chrF_i$ denotes the chrF++ score at beam size $i$.

\item \textbf{Probability gain}: we also extract the sentence generation probabilities by the model and average them over test sentences.
Similar to chrF++, we calculate the probability gain as
$\Delta prob_{1;7} = prob_7 - prob_1$,
where $prob_i$ denotes average generation probability at beam size $i$.
This allows us to estimate the optimality of a decoding strategy from the point of view of the model, independently from a specific quality metric. Since probability gains are strictly positive and vary over orders of magnitude, we apply log-transformation to these values prior to the regression analysis to reduce skewness and heteroscedasticity.
\end{itemize}

Results in Table~\ref{table:regression-gain-nllb} show a low model's fit (adjusted $R^2$) across the board when predicting chrF++ gain. Only source language id and target language resourcedness bear a significant effect on chrF++ gain.
The model predicting probability gain paints a different picture: all explanatory variables have clear and significant predictive power. The coefficients for source languages are positive, meaning that translating from non-English languages yields a higher probability gain compared to English. Since English is arguably the most high-resourced among the languages studied, this could imply that translating not only \textit{into}, but also \textit{from} lower-resource languages (compared to English) may benefit from a wider beam width.
All three target typological properties (genetic distance, MATTR, and head-finality) carry significant positive effects on probability gain, suggesting that narrow beam search may be a worse approximation of optimal model sequence for distant pairs, and that morphologically complex, head-final target languages may benefit from different decoding strategies.

\subsection{Decoder-only Tower+} \label{results:tower}


\begin{table*}[ht]
    \centering
    \footnotesize
    \setlength{\tabcolsep}{6.5pt}
    \begin{tabular}{lrrrrr}
        & \multicolumn{5}{r}{\textit{\textbf{Metric}}: $chrF_7$ ($n=364$)} \\[3pt]
        \textbf{Feature} & $\bm{\beta}$ & \textbf{\textit{F}} & \textbf{\textit{p}-value} & \textbf{\textit{LR}} & \textbf{Adj. \textit{R\textsuperscript{2}}} \\
        \midrule
        \texttt{src\_language} & \textbf{-4.9} & 6.67 & <0.001 & - & 0.0 \\
        \texttt{is\_in\_tower} & \textbf{15.2} & 142 & <0.001 & 218$^*$ & 0.45 \\
        \texttt{tgt:glotcc\_size} & \textbf{6.28} & 102 & <0.001 & 73.2$^*$ & 0.55 \\
        \texttt{is\_same\_script} & \textbf{10.4} & 67.1 & <0.001 & 113$^*$ & 0.67 \\
        \texttt{d\_gen} & -0.04 & 0.004 & 0.95 & 0.5 & 0.67 \\
        \texttt{tgt:mattr} & \textbf{-1.97} & 12.5 & <0.001 & 24.7$^*$ & 0.69 \\
        \texttt{tgt:head\_fin} & \textbf{-3.14} & 29.6 & <0.001 & 29.5$^*$ & 0.71 \\
        \midrule
    \end{tabular}
    \quad
    \begin{tabular}{rrrrr}
        \multicolumn{5}{r}{\textit{\textbf{Metric}}: $\Delta prob_{1;7}$ ($n=364$)} \\[3pt]
        $\bm{\beta}$ & \textbf{\textit{F}} & \textbf{\textit{p}-value} & \textbf{\textit{LR}} & \textbf{Adj. \textit{R\textsuperscript{2}}} \\
        \midrule
        0.15 & 1.58 & 0.15 & - & 0.01 \\
        \textbf{-1.33} & 298 & <0.001 & 302$^*$ & 0.57 \\
        -0.05 & 1.53 & 0.22 & 3.12 & 0.57 \\
        -0.1 & 1.75 & 0.19 & 7.99$^*$ & 0.58 \\
        \textbf{0.16} & 20.2 & <0.001 & 21.7$^*$ & 0.6 \\
        -0.04 & 1.4 & 0.24 & 0.45 & 0.6 \\
        \textbf{0.07} & 4.11 & 0.04 & 4.24$^*$ & 0.6 \\
        \midrule
    \end{tabular}
    \caption{
    Regression results for predicting chrF++ scores at $k=7$ (\textit{left}) and probability gain (\textit{right}) for translations by Tower+ (\textbf{all source languages}). Coefficient values ($\bm{\beta}$) marked in bold are significant ($p<0.05$).}
    \label{table:regression-tower-364}.
\end{table*}

While NLLB-200 is an encoder-decoder Transformer designed specifically for NMT, the current state-of-the-art has shifted towards prompting decoder-only LLMs for translation.
To account for this, we extend our regression experiments to the Tower+ 9B model, using the same 364 language pairs.

\begin{figure}[t]
    \centering
    \includegraphics[width=\linewidth]{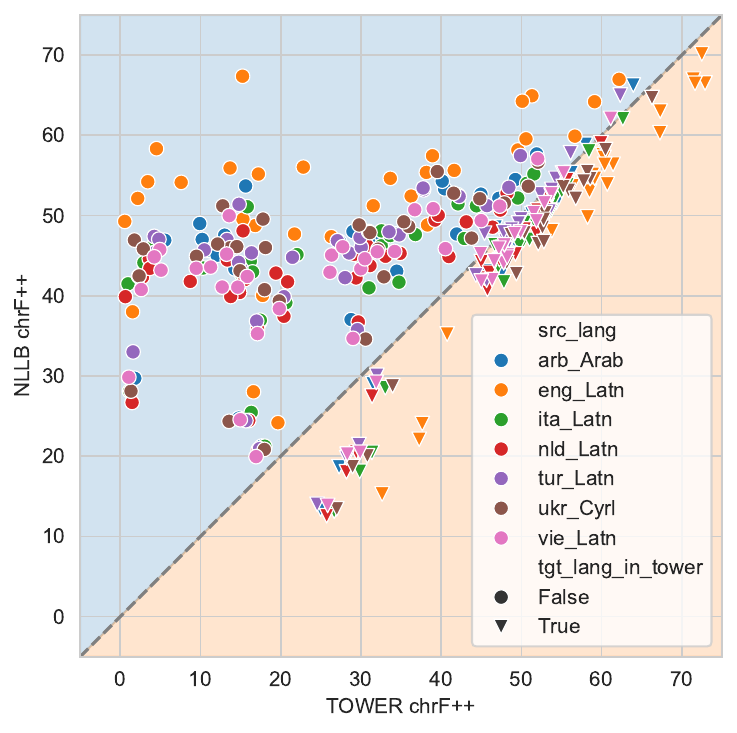}
    \caption{Tower+ 9B chrF++ scores vs. NLLB-200 3.3B chrF++ scores at beam size $k=7$. Each point denotes a language pair and is colored by source language, while $\blacktriangledown$ denotes target languages officially supported by Tower+. The blue and orange shaded regions indicate language pairs for which either NLLB-200 or Tower+ scores are higher, respectively. Sample size is $n = 7 \times 52 = 364$.
    }
    \label{fig:chrf-tower_vs_nllb}
\end{figure}

First, we compare chrF++ scores at beam \mbox{size 7} for translations generated by NLLB-200 and Tower+ (Fig.~\ref{fig:chrf-tower_vs_nllb}).
Tower+ scores higher on the vast majority of target languages it was explicitly trained on (orange region), but underperforms NLLB-200 on \textit{all} target languages outside of Tower's coverage (blue region), reaffirming the relevance of the NLLB model for large-scale MT evaluations including low-resourced languages.

Regression results for predicting $chrF_7$ and probability gain ($\Delta prob_{1;7}$) are shown in Table \ref{table:regression-tower-364}. We use the same predictors as for NLLB-200, but add a simple binary feature \texttt{is\_in\_tower} denoting whether a target language is officially covered by Tower+ or not as per \citet{towerplus}.
When predicting chrF++ scores (Table \ref{table:regression-tower-364}, left),
\CR{almost all variables have a significant effect, barring source-target genetic distance.
The standardized regression coefficients ($\beta$) suggest that disparities in target language resourcedness and model coverage dominate quality variability in the translations produced by Tower+, with the intrinsic typological features of morphology and word order playing a smaller but still significant role.}
The findings for $\Delta prob_{1;7}$ (Table \ref{table:regression-tower-364}, right) are less conclusive, \CR{with fewer significant features than for NLLB-200}.
The directionality of the Tower+ coverage factor ($-1.33$) indicates that non-supported target languages see a larger gain in modeling probability with a larger beam size. \CR{Source language id, target language resourcedness, and source-target script matching do not emerge as significant. Instead, genetic distance and head-finality of the target language have significant predictive power on probability gain and are consistent in their direction and magnitude with the NLLB-200 results.}

\section{Discussion and Conclusion} \label{discussion}

We set out to assess and explain the crosslingual variability of modern MT quality across a broad set of languages, with a focus on word order properties and morphological complexity of the generated (i.e. target) language.
Leveraging the wide coverage FLORES+ MT evaluation dataset and two widely used large pre-trained multilingual MT models (NLLB-200 and Tower+), we generated translations for a variety of language pairs and beam sizes, and evaluated the models' performance and gain thereof via chrF++ and generation probabilities.

Through a set of correlation and regression experiments, we found several language properties to significantly predict quality of the encoder-decoder model (NLLB-200), even after controlling for language resourcedness and script similarity. These properties include source-target typological distances, as well as type/token ratio and head-finality of the target language.

Focusing on inference-time decoding strategy, we uncovered a large variability in chrF++ gains when widening the beam size, calling into question the common practice of using a unique decoding approach for many different language pairs and highlighting the importance of further research in language-specific decoding optimization.
We failed to establish a predictive link between our selected language properties and chrF++ gains by beam size.
However, from the point of view of searching for the model’s highest-probability target sequence,
we did find that languages with more complex morphology and flexible word order benefit more from widening the beam size.
In other words, the standard practice of decoding with a narrow beam search may be particularly suboptimal for these languages.

\CR{Finally, we replicated the regression experiments for the state-of-the-art Tower+ multilingual LLM. When predicting chrF++ score, factors like official language support by the model and resourcedness of the target language seem to dominate the variability of translation quality, but our regression analysis also revealed a strongly significant effect of target morphological complexity and word order. The findings for performance gains were somewhat less conclusive, as fewer language properties seem to have predictive effects on probability delta by beam size. Still, the features that are significant (namely, Tower+ target language coverage, genetic distance, and target head-finality) are consistent in their effects with the encoder-decoder NLLB-200 and demonstrate the significant predictive power of typological properties even when other factors like resourcedness and script emerge as non-significant.}

Promising directions for future work include methods to dynamically set the beam width based on the target language indicated in the user prompt, as well as the evaluation of different strategies beyond beam search which could better account for word order flexibility and other properties of the target language.
We hope that our work will inspire further research on measuring and explaining the intrinsic difficulty of translating and generating different languages in the age of LLMs.

\section*{Limitations} \label{limitations}

While our findings offer insights into how typological features interact with decoding parameters, the scope of our analysis is subject to several constraints.

Firstly, we focus on only two translation models, which enables a controlled and large-scale study, but limits the generalizability of our findings to other models and training paradigms. \CR{Evaluating other (LLM-based) models covering different sets of languages may result in more significant factors in the regression analysis.}

Relying on a surface-level evaluation metric (chrF++) remains problematic. Followup experiments could adopt newer, better metrics as more research is conducted on low-resource languages.

Furthermore, since training language proportions of our evaluated models are not publicly available, we had to estimate language resourcedness through a very rough approximation based on the CommonCrawl dataset.

We also examine only one decoding paradigm --- left-to-right beam search --- with four beam sizes.
Alternative decoding strategies, such as sampling-based or non-autoregressive methods, may interact differently with typological properties, warranting further investigation. 

Finally, our set of typological features is bounded by the availability of data for the languages in our study; for many languages, some fine-grained features remain missing.
Follow-up experiments could at least use more recent versions of the typological distances \citep{khan-etal-2025-uriel, goot-etal-2025-distals}.

\section{Acknowledgements}
We are thankful to the anonymous reviewers for their helpful comments. Vitalii Hirak received funding from the Erasmus Mundus Masters Programme in Language and Communication Technologies, EU grant no. 2019-1508. Jaap Jumelet and Arianna Bisazza were supported by the Talent Programme of the Dutch Research Council (grant VI.Vidi.221C.009).

\bibliography{anthology,biblio,custom}

\appendix

\section{Target Languages} \label{app:tgt_langs}

\begin{table*}[ht!]
    \centering
    \footnotesize
    \setlength{\tabcolsep}{10pt}
    \begin{tabular}{lllll}
        \toprule
        Acehnese (Arabic script) & \textbf{Armenian} & \textbf{Portuguese (Brazilian)} \\
        Acehnese (Latin script) & Igbo & Dari \\
        Mesopotamian Arabic & Ilocano & Ayacucho Quechua \\
        \textbf{Afrikaans} & \textbf{Indonesian} & \textbf{Romanian} \\
        Amharic & \textbf{Icelandic} & Rundi \\
        \textbf{Modern Standard Arabic} & \textbf{Italian} & \textbf{Russian} \\
        Moroccan Arabic & \textbf{Japanese} & Sango \\
        Egyptian Arabic & Jingpho & \textbf{Sanskrit} \\
        Assamese & Kamba & \textbf{Slovak} \\
        Central Aymara & Kannada & \textbf{Slovenian} \\
        South Azerbaijani & Georgian & Samoan \\
        Bashkir & \textbf{Kazakh} & Shona \\
        Bambara & Halh Mongolian & Somali \\
        Balinese & Khmer (Central) & Southern Sotho \\
        \textbf{Belarusian} & Kikuyu & \textbf{Spanish (Latin American)} \\
        Bhojpuri & Kyrgyz & \textbf{Serbian} \\
        Lhasa Tibetan & Northern Kurdish & Swati \\
        \textbf{Bulgarian} & Central Kanuri (Arabic script) & Sundanese \\
        \textbf{Catalan} & Central Kanuri (Latin script) & \textbf{Swedish} \\
        \textbf{Czech} & \textbf{Korean} & Swahili \\
        Central Kurdish & Lao & \textbf{Tamil} \\
        \textbf{Mandarin Chinese (Standard Beijing)} & \textbf{Lithuanian} & Tamasheq (Latin script) \\
        \textbf{Mandarin Chinese (Taiwanese)} & Ganda & Tamasheq (Tifinagh script) \\
        \textbf{Welsh} & Luo & Tatar \\
        \textbf{Danish} & Mizo & \textbf{Telugu} \\
        \textbf{German} & Maithili & Tajik \\
        \textbf{Estonian} & Malayalam & Thai \\
        \textbf{Greek} & \textbf{Marathi} & Tigrinya \\
        \textbf{English} & Minangkabau (Latin script) & Tswana \\
        \textbf{Basque} & Meitei (Manipuri, Bengali script) & Turkmen \\
        Ewe & Mossi & \textbf{Turkish} \\
        Fijian & Maori & \textbf{Uyghur} \\
        \textbf{Finnish} & Burmese & \textbf{Ukrainian} \\
        \textbf{French} & \textbf{Dutch} & \textbf{Urdu} \\
        \textbf{Scottish Gaelic} & Nepali & \textbf{Vietnamese} \\
        \textbf{Irish} & Nuer & Waray \\
        \textbf{Galician} & Odia & \textbf{Wolof} \\
        Hausa & Pangasinan & Xhosa \\
        \textbf{Hebrew} & Eastern Panjabi & Yoruba \\
        \textbf{Hindi} & \textbf{Western Persian} & Yue Chinese (Hong Kong Cantonese) \\
        \textbf{Croatian} & Plateau Malagasy & Zulu \\
        \textbf{Hungarian} & \textbf{Polish} &  \\
        \bottomrule
    \end{tabular}
    \caption{125 target languages used for our \textit{correlation} experiments (including English). Languages \textbf{in bold} are used in \textit{regression} experiments (53 languages).}
    \label{table:tgt_langs}
\end{table*}

Table \ref{table:tgt_langs} provides the list of target languages used in our experiments.

\section{More Details about the Language Properties} \label{app:properties}

Besides language resourcedness and coarse-grained source-target distances, we focus on features of morphological complexity and word order of the language being generated.
With respect to \textbf{morphology}, current subword segmentation methods have been proved less effective for more morphologically complex languages \citep{park-etal-2021-morphology}.
\textbf{Word order} flexibility is linked to entropy and potentially less predictable sequences. 
Moreover, word order properties can determine how information is distributed within the sentence \citep{maurits2010some,trujillo2024information},\footnote{Some languages tend to place less predictable words and phrases in the first half of the utterance (e.g. German, Japanese), whereas others follow the opposite pattern (e.g. English, Mandarin).} which in turn could affect the optimal decoding strategy of a given target language.

\subsection{Precomputed Distances}
Following previous work on NMT \cite{sarti-etal-2022-divemt} and cross-lingual transfer \cite{lin-etal-2019-choosing}, we take advantage of the URIEL typological database \cite{littell-etal-2017-uriel} containing vector representations of numerous languages drawn from typological, geographical, and phylogenetic databases. Using the accompanying \texttt{lang2vec} library\footnote{\href{https://github.com/antonisa/lang2vec}{\tt github.com/antonisa/lang2vec}.}, we query six types of precomputed distances between each FLORES+ language and seven source languages: \textit{genetic}, \textit{geographic}, \textit{syntactic}, \textit{inventory}, \textit{phonological}, and \textit{overall features}.

\subsection{Morphological Complexity Measures}

To estimate morphological complexity of a language, we make use of eight publicly available precalculated continuous measures from \citet{ccoltekin2023complexity}, outlined below. The measures were computed on the basis of Universal Dependencies and are available for 33 languages of the FLORES+ dataset.

\paragraph{Type/Token Ratio (TTR)} Number of unique word types in a text divided by the total number of word tokens in a text \cite{chotlos1944iv, covington2010cutting}. Since TTR lies in the range of $[0;1]$, languages where this measure is closer to 1 will have a higher number of unique word forms in part motivated by inflectional morphology, which we expect to negatively affect translation difficulty.

\paragraph{Information in Word Structure (WS)} Compares the information content (i.e. entropy) of the original text with its compressed version \cite{juola1998measuring}. The expectation here is that languages with more complex morphology will have worse compression ratios.

\paragraph{Word and Lemma Entropy (WH, LH)} Word entropy is calculated on the basis of word frequency distribution of a text, with morphologically complex languages having higher word entropy \cite{bentz-etal-2016-comparison}. \citet{ccoltekin2023complexity} additionally calculate the entropy of lemmas. Since lemmas do not include inflectional markers, a high degree of lemma entropy would then point at more derivational morphology and compounding.

\paragraph{Mean Size of Paradigm (MSP)} Calculated by dividing the number of word forms in a text by the number of lemmas \cite{xanthos2011role}. Languages with richer inflectional morphology are expected to have a higher number of paradigm cells, reflected by the MSP value.

\paragraph{Inflectional Synthesis (IS)} Maximum number of inflection categories that can be expressed by a standalone verb.

\paragraph{Morphological Feature Entropy (MFH)} Reflects the usage of morphological features and their values. Languages with a higher number of approximately uniformly used grammatical cases will have higher entropy values, indicating more intricate inflectional morphology.

\paragraph{Inflection Accuracy (IA)} Accuracy of an ML model on the task of predicting inflected forms of words given their lemmas and grammatical features. The intuition for IA is that if a language has complex morphology, inflection accuracy on a hold-out test set will be low. Thus, for the sake of consistency with the rest of the measures, we report negative inflection accuracy (\texttt{-ia}).

\paragraph{TTR Measured on FLORES+} We leverage the LexicalRichness Python module \cite{lex} to calculate three TTR measures on the data for the languages covered by the \texttt{dev} split of FLORES+: \textit{TTR}, \textit{Root Type/Token Ratio} (RTTR) \cite{guiraud1959problemes}, and \textit{Moving Average Type/Token Ratio} (MATTR) \cite{covington2010cutting}.

\subsection{Gradient Word Order Measures}

We include a number of gradient word order measures proposed in \citet{levshina2019token} and \citet{levshina2023we}.

\paragraph{Average Word Order Entropy of Dependents and Codependents} Entropy of different word order patterns of \textit{dependencies} (e.g. verb-subject and noun-adposition relations) and \textit{codependencies} (e.g. subject and object of the same verb).

\paragraph{Proportion of Subject-Object Order} Proportions based on the frequencies of phrases where subject comes before object. Proportions closer to 1 indicate strong preference of a language towards either order of subject and object, while proportions closer to 0.5 mean that a language tends to use the two orders interchangeably. \CR{For consistency with other metrics, we report results with \textit{negative} SO order proportion {\tt -SO\_prop}, such that higher {\tt -SO\_prop} values reflect \textit{lower} predictability of word order.}

\paragraph{Percentage of Head-Final Phrases} Approximates the preference of a language towards head-initial or head-final phrases. Values closer to 1 indicate a larger prevalence of head-finality.

\section{Tower+ Translation Prompt} \label{app:prompt}

\texttt{Translate the following \{src\_lang\} source text to \{tgt\_lang\}:\textbackslash n\{src\_lang\}: \{src\_text\}\textbackslash n\{tgt\_lang\}: }

\section{Overall Results with Additional Metrics} 
\label{app:other metrics}

\begin{figure}[h]
    \centering
    \includegraphics[width=\linewidth]{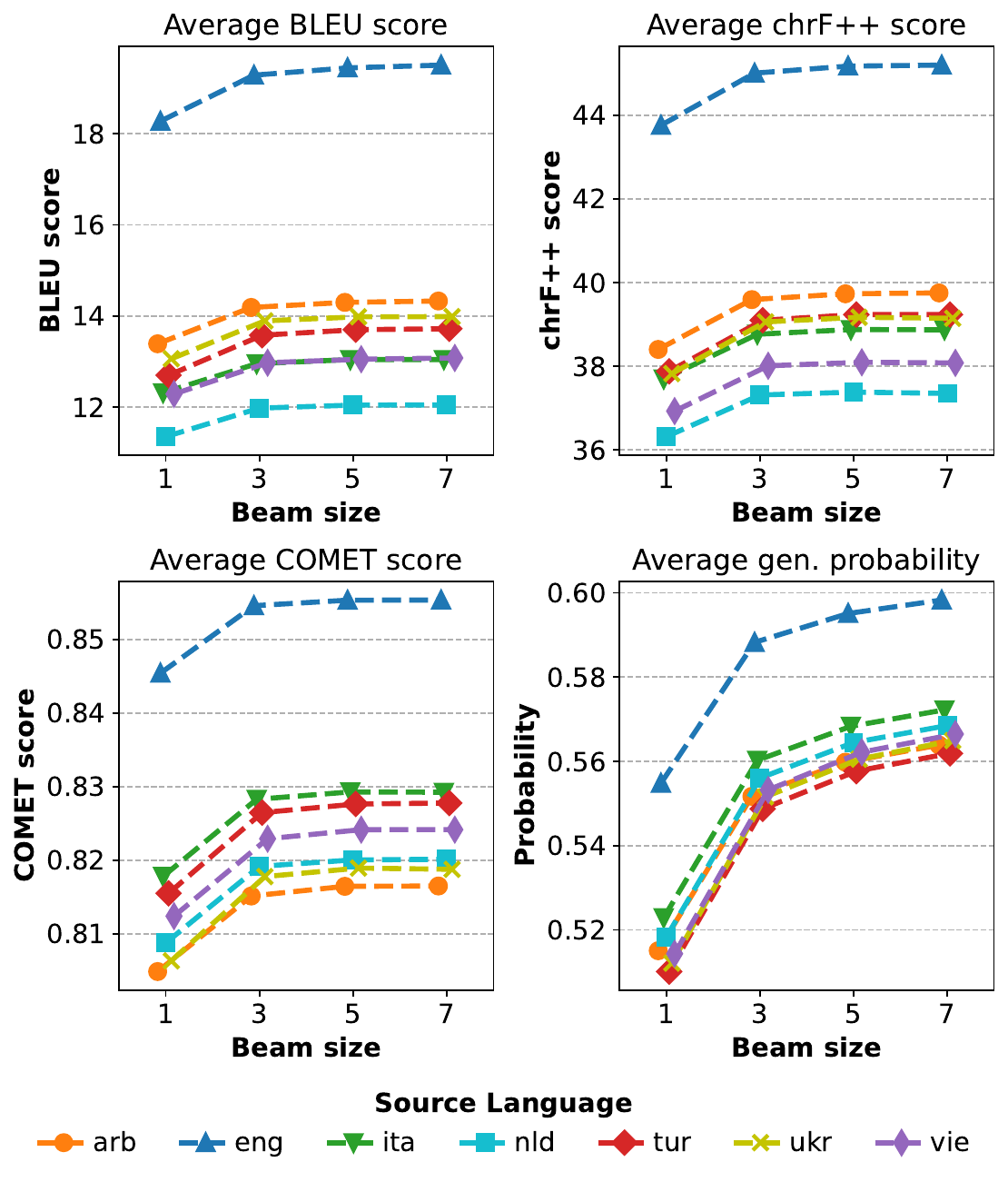}
    \caption{NLLB-200 translation performance measured at four beam sizes via translation quality metrics (BLEU, chrF++, COMET) and output sequence generation probabilities. Performance is averaged across 124 target languages for individual source languages: Arabic, English, Italian, Dutch, Turkish, Ukrainian, and Vietnamese.}
    \label{fig:beamsize}
\end{figure}

Figure \ref{fig:beamsize} shows NLLB-200 translation quality results with different metrics (BLEU, chrF++, COMET), as well as sentence generation probabilities, as a function of beam size.
Scores are averaged across 124 target languages. 
Trends by beam size are very similar across metrics. 
While a wider beam generally improves translation quality and the model's confidence during output sequence generation, regardless of the source language, it is not clear whether the \textit{degree} of this improvement varies depending on the typological properties of target languages, motivating our correlation and regression analyses.

\section{Effect of Head-Finality and MATTR} \label{app:regplots}

\begin{figure}[h]
    \centering
    \includegraphics[width=\linewidth]{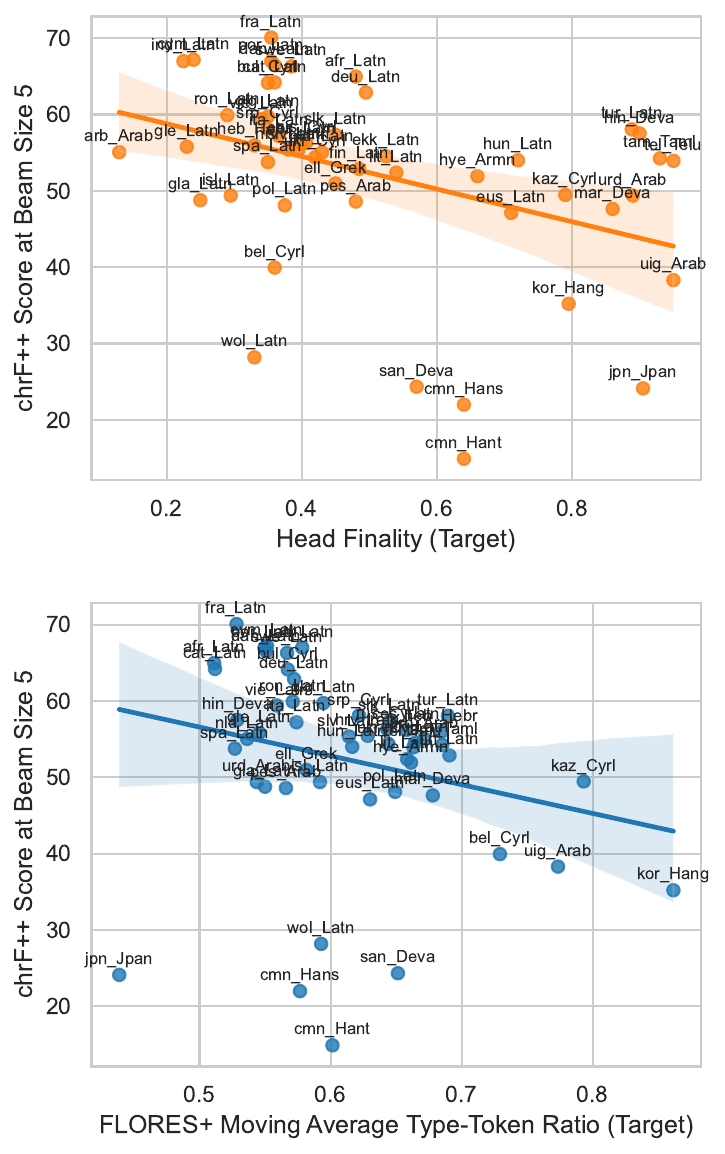}
    \caption{Target language head-finality (top) and moving average type-token ratio (bottom) vs. chrF++ scores at beam size 5. Data for translations by NLLB-200 from English into 52 target languages.}
    \label{fig:regplots}
\end{figure}

Figure \ref{fig:regplots} provides a closer look at the negative effect of target language head finality and moving average type-token ratio on NLLB-200 chrF++ scores when translating from English.

\section{Additional Regression Results} 

\begin{table*}[ht]
    \centering
    \footnotesize
    \setlength{\tabcolsep}{4pt}
    \begin{tabular}{lrlrrrr}
        & \multicolumn{5}{c}{{\textit{Metric}: \textbf{chrF++ Gain} ($n=364$)}} \\[3pt]
        \textbf{Feature} & $\bm{\beta}$ & \textbf{\textit{F}-statistic} & \textbf{\textit{p}-value} & \textbf{\textit{LR}} & \textbf{Adj. \textit{R\textsuperscript{2}}} \\
        \midrule
        \texttt{src\_language} & -- & 4.40 & $<$0.001 & -- & 0.05 \\
        {\scriptsize\hspace{0.5cm}\textit{Arabic}} & {\scriptsize\textbf{-0.26}} \\
        {\scriptsize\hspace{0.5cm}\textit{Dutch}} & {\scriptsize\textbf{-0.60}}\\
        {\scriptsize\hspace{0.5cm}\textit{Italian}} & {\scriptsize\textbf{-0.37}}\\
        {\scriptsize\hspace{0.5cm}\textit{Turkish}} &  {\scriptsize-0.12}\\
        {\scriptsize\hspace{0.5cm}\textit{Ukrainian}} & {\scriptsize-0.11}\\
        {\scriptsize\hspace{0.5cm}\textit{Vietnamese}} & {\scriptsize\textbf{-0.36}}\\[2pt]
        \texttt{tgt:glotcc\_size} & \textbf{0.15} & 17.2 & $<$0.001 & 14.2$^*$ & 0.09 \\
        \texttt{is\_same\_script} & -0.02 & 0.06 & 0.809 & 1.72 & 0.09 \\
        \texttt{d\_gen} & 0.05 & 1.48 & 0.224 & 2.00 & 0.09 \\
        \texttt{tgt:mattr} & 0.02 & 0.35 & 0.555 & 1.04 & 0.09 \\
        \texttt{tgt:head\_fin} & 0.06 & 2.38 & 0.124 & 2.45 & 0.09 \\
        \bottomrule
    \end{tabular}
    \quad
    \begin{tabular}{rrrrr}
    \multicolumn{5}{c}{{\textit{Metric}: \textbf{Probability Gain} ($n=364$)}} \\[3pt]
     $\bm{\beta}$ & \textbf{\textit{F}-statistic} & \textbf{\textit{p}-value} & \textbf{\textit{LR}} & \textbf{Adj. \textit{R\textsuperscript{2}}} \\
    \midrule
    -- & 3.6 & 0.002 & -- & 0.01 \\
    {\scriptsize 0.09 }& \\
    {\scriptsize\textbf{0.19}}& \\
    {\scriptsize\textbf{0.19}}& \\
    {\scriptsize\textbf{0.19}}& \\
    {\scriptsize\textbf{0.21}}& \\
    {\scriptsize\textbf{0.19}}& \\[2pt]
    \textbf{-0.21} & 180.8 & $<$0.001 & 141.8$^*$ & 0.33 \\
    \textbf{-0.08} & 4.5 & 0.03 & 37.2$^*$ & 0.39 \\
    \textbf{0.09} & 26.9 & $<$0.001 & 29.7$^*$ & 0.44 \\
   \textbf{ 0.05} & 10.6 & 0.001 & 23.1$^*$ & 0.47 \\
    \textbf{0.12} & 46.3 & $<$0.001 & 45.0$^*$ & 0.53 \\
    \bottomrule
    \end{tabular}
    \caption{
    Regression results for chrF++ gain (\textit{left}) and probability gain (\textit{right}) for NLLB-200 translations (\textbf{all source languages}); model is fitted on the results of all source languages. Coefficient values ($\bm{\beta}$) marked in bold are significant ($p<0.05$).
    }
    \label{table:regression-gain-nllb-src_langs}
\end{table*}

\subsection{NLLB-200} 
\label{app:regression-gain-nllb-src_lang}

Table \ref{table:regression-gain-nllb-src_langs} summarizes regression results for predicting chrF++ and probability gain for translations by NLLB-200. Here we indicate the coefficient values $\bm{\beta}$ for each source language, with English as a reference value.

\subsection{Tower+: chrF++ gain} \label{app:regression-chrfgain-tower}

Table \ref{table:regression-tower-chrf_gain-364} showcases regression results for predicting chrF++ gain of Tower+ translations.


\begin{table}[ht]
    \centering
    \footnotesize
    \setlength{\tabcolsep}{3pt}
    \begin{tabular}{lrrrrr}
        & \multicolumn{5}{r}{\textit{\textbf{Metric}}: $\Delta chrF_{1;7}$ ($n=364$)} \\[3pt]
        \textbf{Feature} & $\bm{\beta}$ & \textbf{\textit{F}} & \textbf{\textit{p}-value} & \textbf{\textit{LR}} & \textbf{Adj. \textit{R\textsuperscript{2}}} \\
        \midrule
        \texttt{src\_language} & 0.0 & 1.74 & 0.11 & - & 0.01 \\
        \texttt{is\_in\_tower} & -0.48 & 1.35 & 0.25 & 0.74 & 0.01 \\
        \texttt{tgt:glotcc\_size} & \textbf{0.66} & 11.1 & <0.001 & 13.61$^*$ & 0.04 \\
        \texttt{is\_same\_script} & -0.6 & 2.19 & 0.14 & 0.29 & 0.04 \\
        \texttt{d\_gen} & 0.11 & 0.38 & 0.54 & 0.12 & 0.04 \\
        \texttt{tgt:mattr} & -0.11 & 0.36 & 0.55 & 1.77 & 0.04 \\
        \texttt{tgt:head\_fin} & \textbf{-0.47} & 6.42 & 0.01 & 6.6$^*$ & 0.05 \\
        \midrule
    \end{tabular}
    \quad
    \caption{
    Regression results for predicting chrF++ gain for translations generated by \textbf{Tower+} (\textbf{all src langs}).}
    \label{table:regression-tower-chrf_gain-364}
\end{table}

\break

\section{Correlation Results for All Source Languages} \label{app:correlations-all}

\begin{figure*}
    \centering
    \includegraphics[width=1\linewidth]{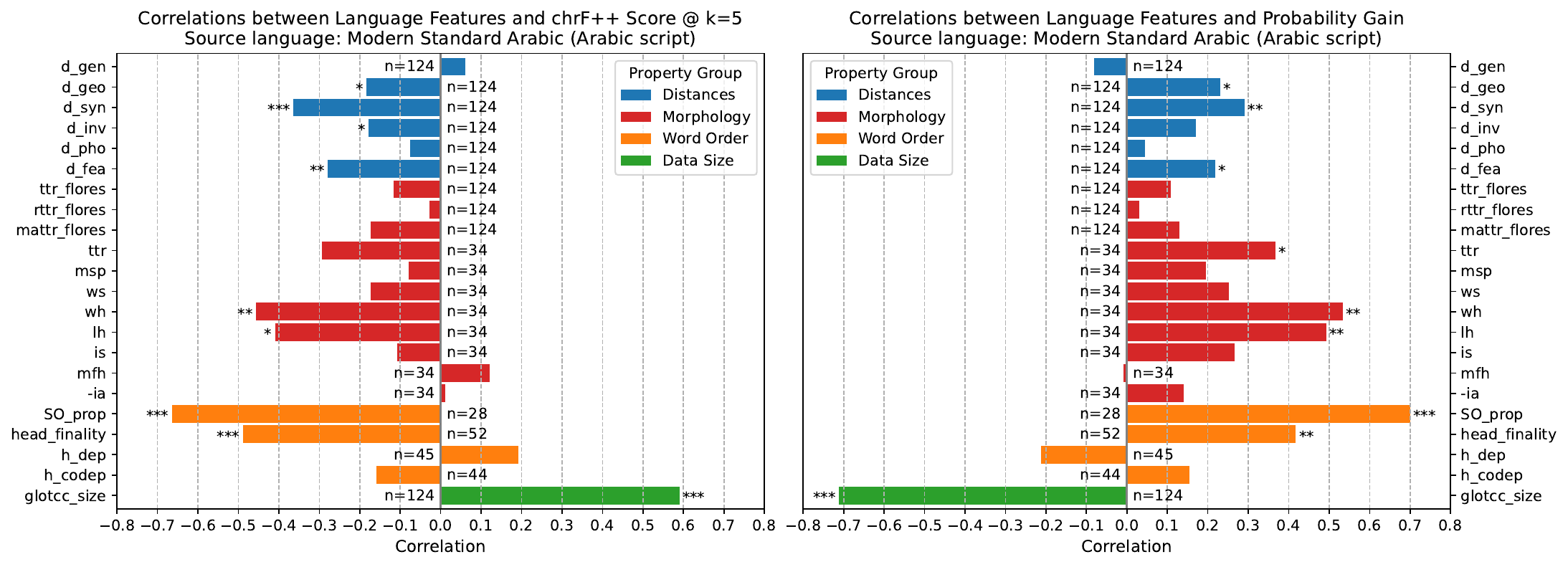}
    \caption{Spearman correlations between continuous language properties and chrF++ scores at beam size $k=5$ (left) and probability gain for beam size $k=7$ (right). Source language: \textbf{Arabic}.}
    \label{fig:corrs-arb}
\end{figure*}

\begin{figure*}
    \centering
    \includegraphics[width=1\linewidth]{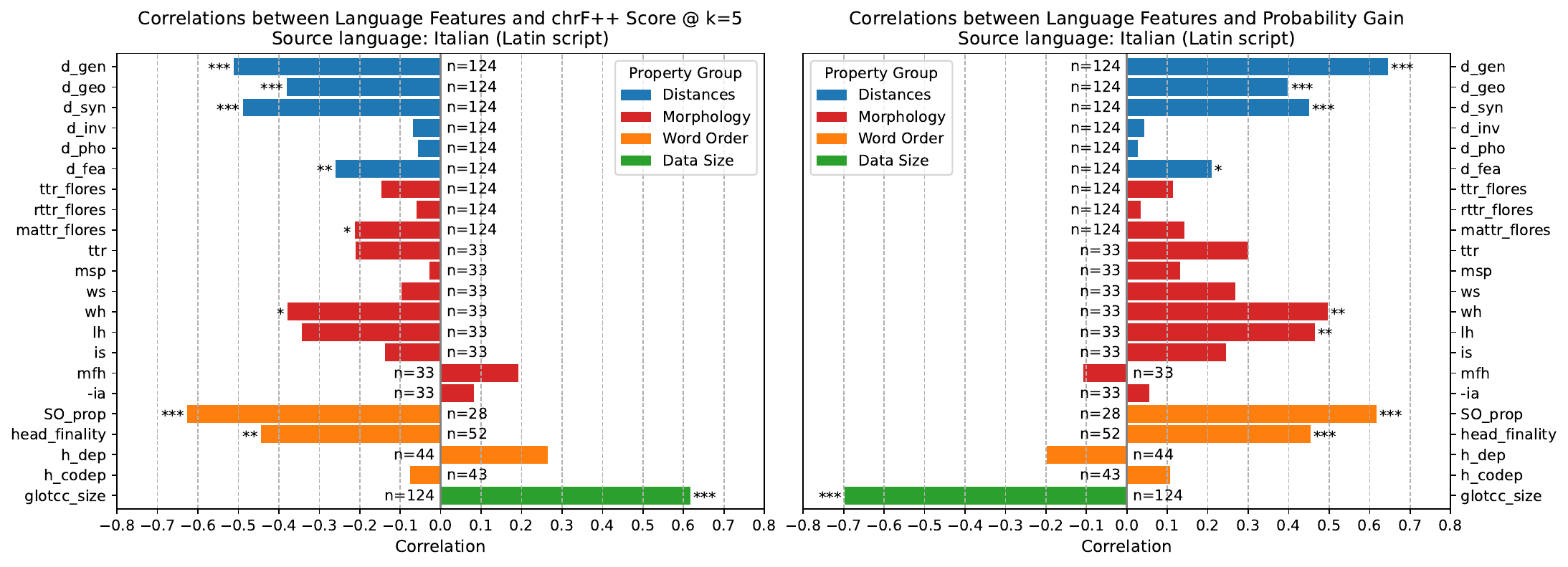}
    \caption{Spearman correlations between continuous language properties and chrF++ scores at beam size $k=5$ (left) and probability gain for beam size $k=7$ (right). Source language: \textbf{Italian}.}
    \label{fig:corrs-ita}
\end{figure*}

\begin{figure*}
    \centering
    \includegraphics[width=1\linewidth]{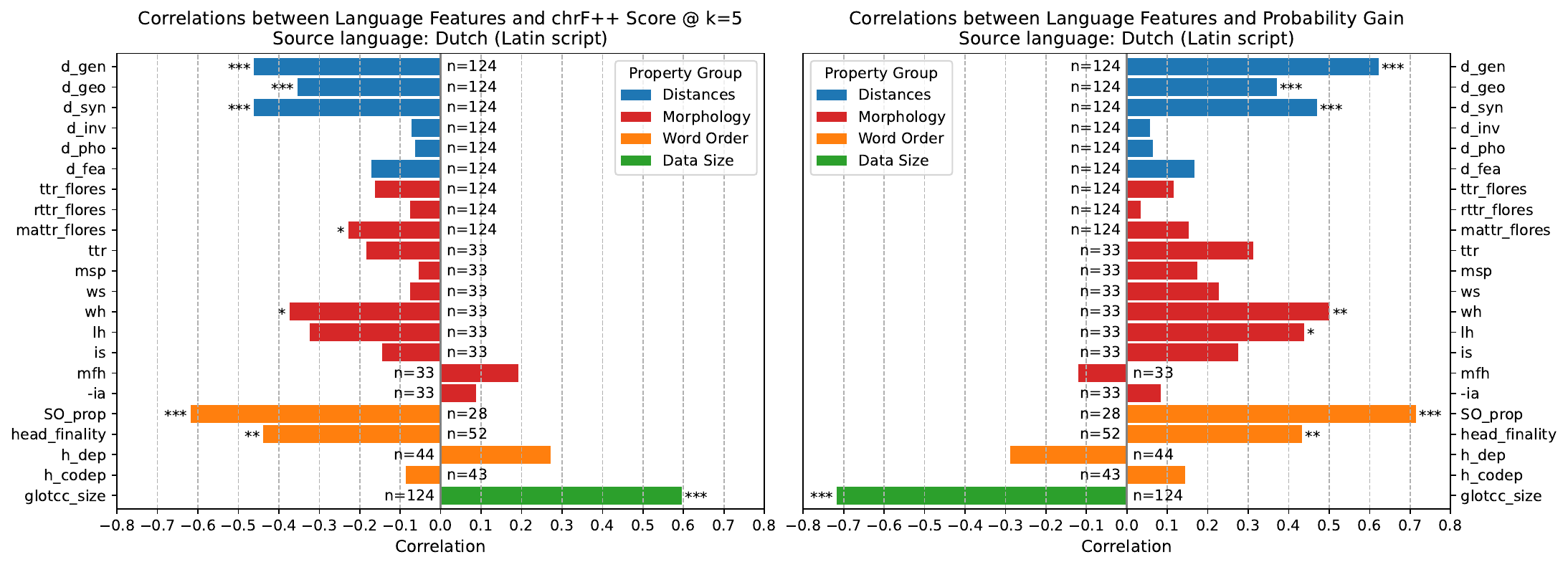}
    \caption{Spearman correlations between continuous language properties and chrF++ scores at beam size $k=5$ (left) and probability gain for beam size $k=7$ (right). Source language: \textbf{Dutch}.}
    \label{fig:corrs-nld}
\end{figure*}

\begin{figure*}
    \centering
    \includegraphics[width=1\linewidth]{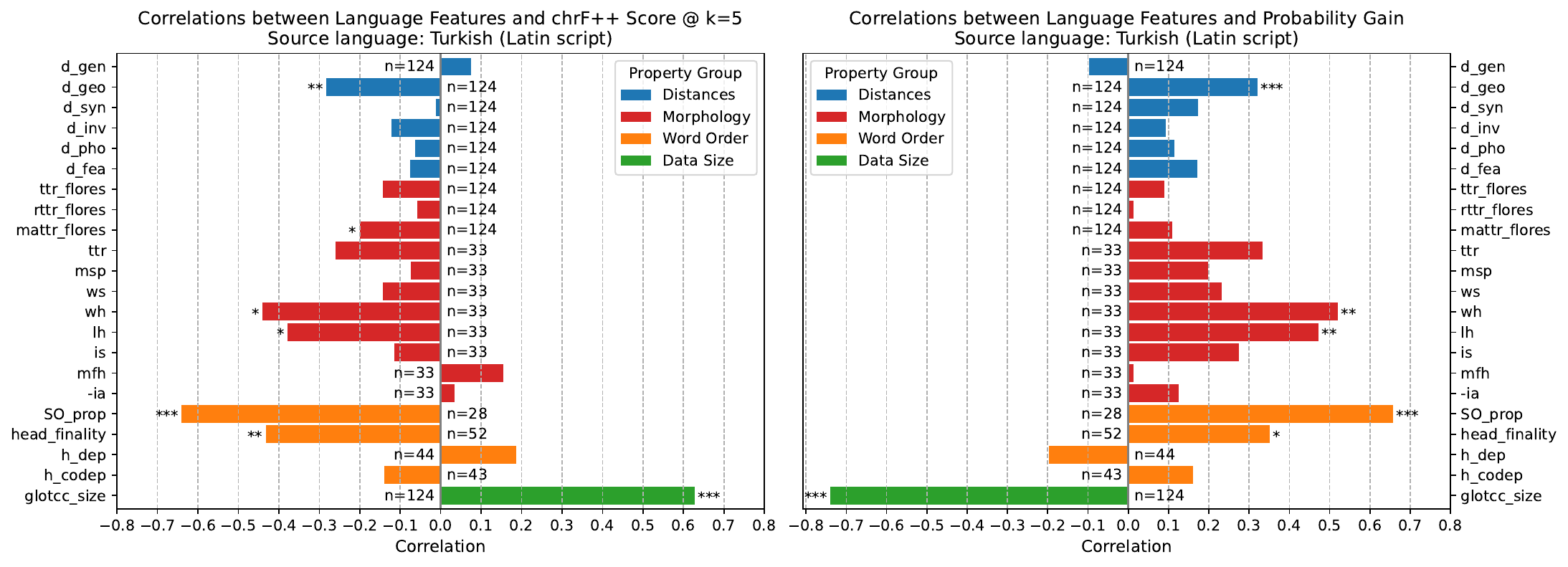}
    \caption{Spearman correlations between continuous language properties and chrF++ scores at beam size $k=5$ (left) and probability gain for beam size $k=7$ (right). Source language: \textbf{Turkish}.}
    \label{fig:corrs-tur}
\end{figure*}

\begin{figure*}
    \centering
    \includegraphics[width=1\linewidth]{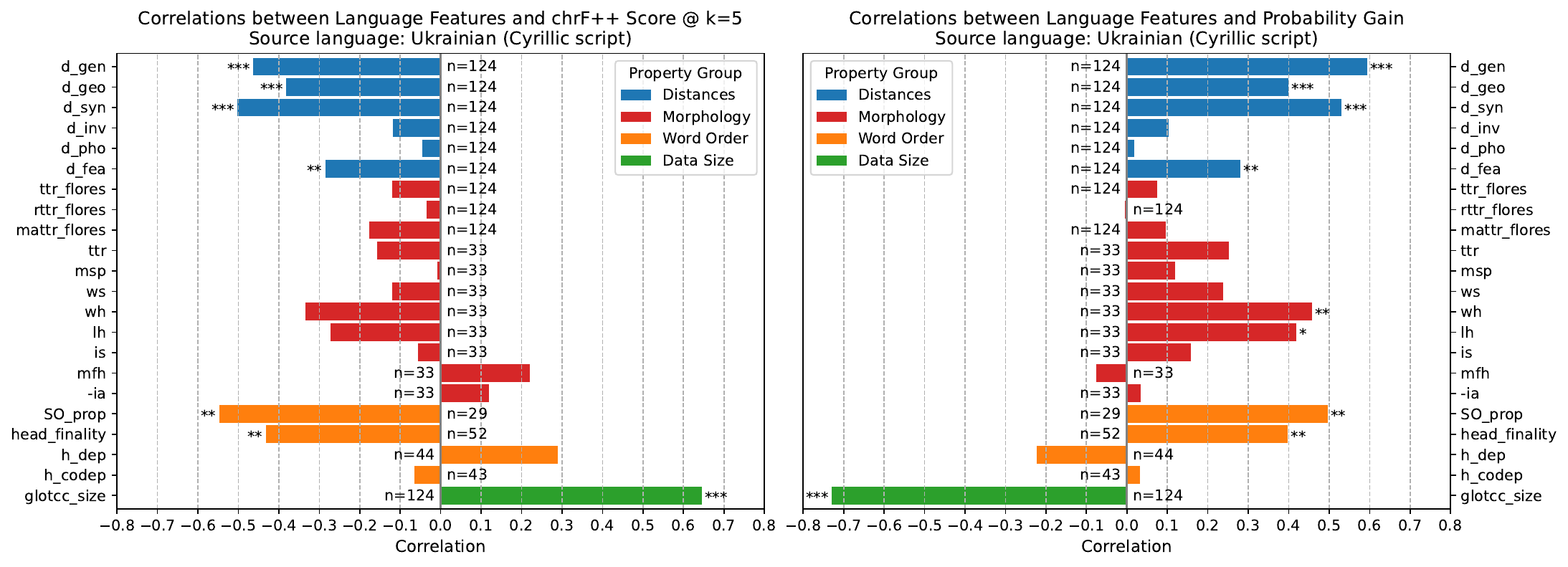}
    \caption{Spearman correlations between continuous language properties and chrF++ scores at beam size $k=5$ (left) and probability gain for beam size $k=7$ (right). Source language: \textbf{Ukrainian}.}
    \label{fig:corrs-ukr}
\end{figure*}

\begin{figure*}
    \centering
    \includegraphics[width=1\linewidth]{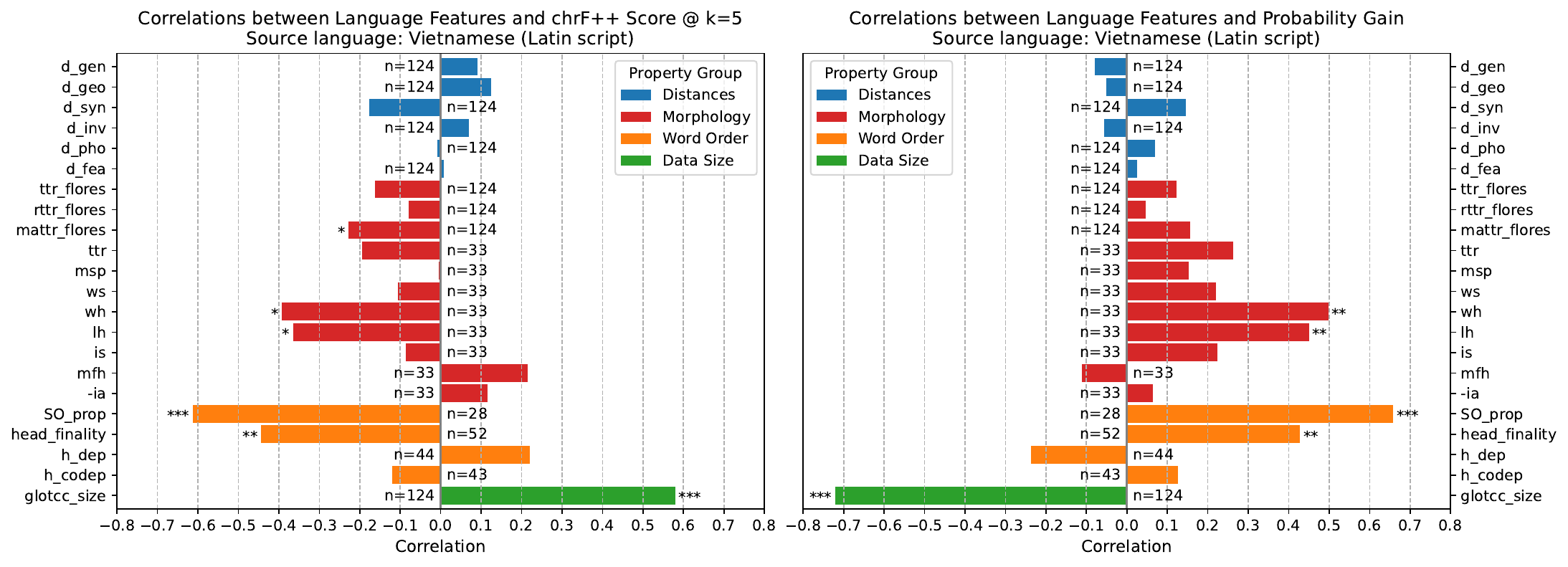}
    \caption{Spearman correlations between continuous language properties and chrF++ scores at beam size $k=5$ (left) and probability gain for beam size $k=7$ (right). Source language: \textbf{Vietnamese}.}
    \label{fig:corrs-vie}
\end{figure*}

Here we include Spearman correlation results for chrF++ scores at beam size $k=5$ and generation probability gains for NLLB-200 translation from six source languages:

\begin{itemize}
    \item Arabic: Figure \ref{fig:corrs-arb}
    \item Italian: Figure \ref{fig:corrs-ita}
    \item Dutch: Figure \ref{fig:corrs-nld}
    \item Turkish: Figure \ref{fig:corrs-tur}
    \item Ukrainian: Figure \ref{fig:corrs-ukr}
    \item Vietnamese: Figure \ref{fig:corrs-vie}
\end{itemize}

\end{document}